\documentclass[10pt,twocolumn,letterpaper]{article}
\usepackage[pagenumbers]{cvpr} %
\usepackage{graphics,graphicx,caption,float,subcaption,booktabs,xcolor,multirow,array,color,ifthen,tabu,colortbl,dblfloatfix,url,relsize,algorithm,algorithmic,amssymb,xspace,nicefrac,microtype,amsmath,amsfonts,bm,wrapfig,makecell,enumitem}
\usepackage[pagebackref=true,breaklinks=true,colorlinks=true,bookmarks=false]{hyperref}
\usepackage[capitalize]{cleveref}
\usepackage[english]{babel}
\usepackage[accsupp]{axessibility}

\crefname{section}{Sec.}{Secs.}
\Crefname{section}{Section}{Sections}
\Crefname{table}{Table}{Tables}
\crefname{table}{Tab.}{Tabs.}

\newcommand\blfootnote[1]{%
  \begingroup
  \renewcommand\thefootnote{}\footnote{#1}%
  \addtocounter{footnote}{-1}%
  \endgroup
}

\newcommand{\paragraphc}[1]{\vspace{0.3em}\noindent\textbf{#1}}

\newcommand{\pagelink}{https://navigation-locomotion.github.io}

\newcommand{\algo}{VP-Nav\xspace}

\def\cvprPaperID{19} %
\def\confName{CVPR}
\def\confYear{2022}

\begin{document}

\title{Coupling Vision and Proprioception for Navigation of Legged Robots}

\author{Zipeng Fu\textsuperscript{*1}\;\;\; Ashish Kumar\textsuperscript{*2}\;\;\; Ananye Agarwal\textsuperscript{1}\;\;\; Haozhi Qi\textsuperscript{2}\;\;\; Jitendra Malik\textsuperscript{2}\;\;\; Deepak Pathak\textsuperscript{1}\vspace{2mm}\\
\textsuperscript{1}Carnegie Mellon University \;\;\; \textsuperscript{2}UC Berkeley
}

\maketitle

\begin{abstract}
\vspace{-0.2cm}
We exploit the complementary strengths of vision and proprioception to develop a point-goal navigation system for legged robots, called \algo. Legged systems are capable of traversing more complex terrain than wheeled robots, but to fully utilize this capability, we need a high-level path planner in the navigation system to be aware of the walking capabilities of the low-level locomotion policy in varying environments. We achieve this by using proprioceptive feedback to ensure the safety of the planned path by sensing unexpected obstacles like glass walls, terrain properties like slipperiness or softness of the ground and robot properties like extra payload that are likely missed by vision. The navigation system uses onboard cameras to generate an occupancy map and a corresponding cost map to reach the goal. A fast marching planner then generates a target path. A velocity command generator takes this as input to generate the desired velocity for the walking policy. A safety advisor module adds sensed unexpected obstacles to the occupancy map and environment-determined speed limits to the velocity command generator. We show superior performance compared to wheeled robot baselines, and ablation studies which have disjoint high-level planning and low-level control. We also show the real-world deployment of \algo on a quadruped robot with onboard sensors and computation. Videos at~\url{\pagelink}
\end{abstract}

\vspace{-1.2em}
\section{Introduction}
\vspace{-1.2em}
\label{sec:intro}

\blfootnote{*equal contribution}

Gibson has famously remarked, ``we see in order to move and we move in order to see.'' Although, it would be more accurate to say that we \textit{see} and \textit{feel} in order to move. Vision and proprioception are complementary senses. Vision is a distance sense, which allows us to avoid static and dynamic obstacles. However, vision is slow and cannot directly sense physical properties of terrains such as softness vs. hardness, smooth vs. rough. Proprioception (knowledge of agent's own body like joint angles, body orientation, foot contacts, etc.) is fast and gives a direct measurement of physical environment characteristics. In this paper, we will focus on exploiting the complementary strengths of vision and proprioception for navigation of legged robots. The goal is to train a legged robot by developing both low-level control of its motor joints to walk on terrains (i.e., locomotion) as well as high-level path planning to reach certain goal locations by autonomously avoiding any obstacles along the way (i.e., navigation).

\begin{figure}[t]
\centering
\includegraphics[width=\linewidth]{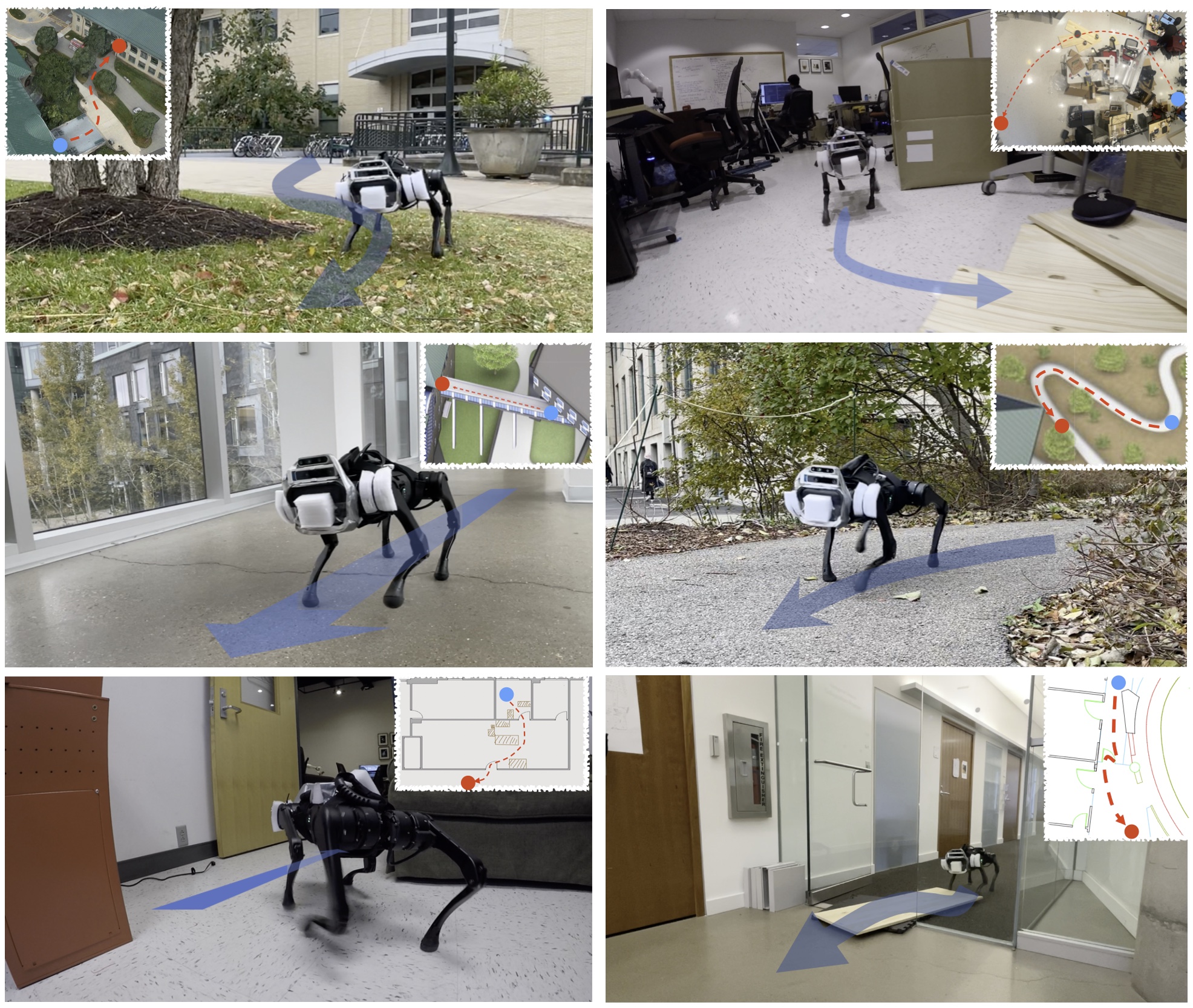}
\vspace{-2em}
\caption{\small Example deployment scenarios for our proposed point goal navigation system for legged robots. The varying terrains on the way to the goal require the planner to be aware of the robot's locomotion capabilities. The proprioceptive coupling between the locomotion controller and the navigation planner allow the robot to sense properties of the environment which the vision might miss (slippery terrain, glass obstacle, etc.).}
\vspace{-1.3em}
\label{fig:teaserreal}
\end{figure}

\begin{figure*}[t]
    \vspace{-1.0cm}
    \centering

  \includegraphics[width=1.0\linewidth,height=0.44\linewidth]{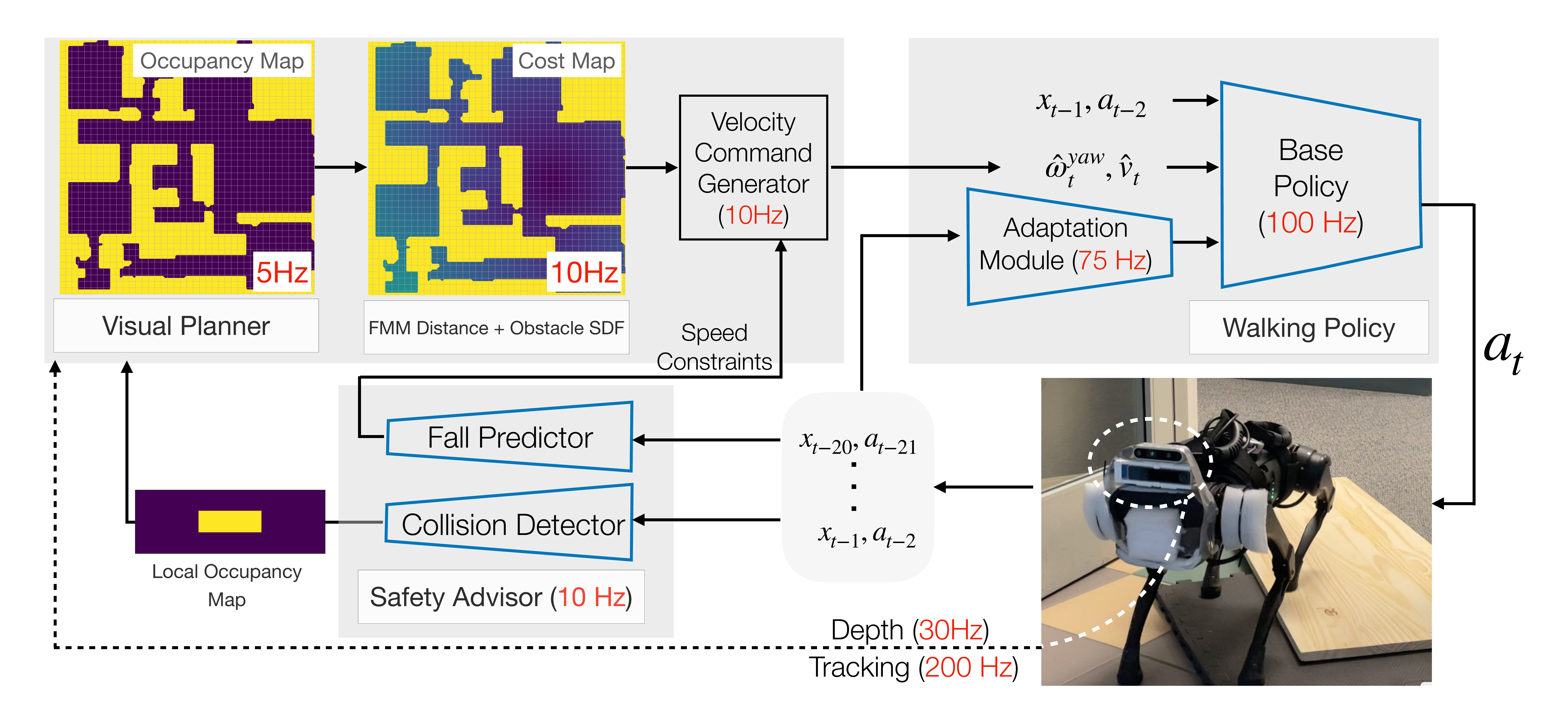}
  \vspace{-3em}
  \caption{\small Our navigation system (\algo) consists of a velocity-conditioned walking policy, a safety advisor module, and a planning module. The velocity-conditioned walking policy takes as input the command velocity and the proprioceptive robot state to output the actions needed to walk in a variety of complex settings. Once we have learned the walking policy in simulation, we then train a Safety Advisor Module, also in simulation, which  estimates the safety constraints of the walking policy. It uses proprioception to estimate two probabilities: (1) if the robot is in collision, (2) if the robot is about to fall, which are used to update the map and velocity estimates to walk safely in its environment. The planner uses on board cameras to compute a navigation cost map for an input point goal and takes in the safety constraints from the Safety Advisor module to compute desired walking velocity and direction. All modules run asynchronously onboard of the robot. 
  }
    \label{fig:method}
    \vspace{-1em}
\end{figure*}

\paragraphc{Locomotion and Navigation:}
Traditionally, locomotion and navigation are studied as separate problems and then put together on a robot as individual modules \cite{wooden2010autonomous, truong2020learning, li2021planning, hoeller2021learning}. However, to truly support dynamic goal reaching in complex terrains, the planner should know about the walking ability of the robot in different terrains. For instance, a robot navigating to a goal through a slippery patch may either lower its walking speed or walk around it altogether depending on its locomotion ability. To facilitate such communication between high-level and low-level, prior works generally infer a cost map for the planner from an onboard vision sensor which is only capable of detecting clearly visible obstacles and regions that are hard to traverse, e.g. steps and ramps~\cite{wooden2010autonomous,wermelinger2016navigation,chilian2009stereo,yang2021real}.  However, it is extremely challenging to predict several other terrain properties from vision like how slippery, uneven, granular or deformable the surface is. These directly affect the walking robot's ability to follow the plan. Furthermore, the environment could also contain obstacles that are invisible to a vision-only planner as shown in Figure~\ref{fig:teaserreal} and Figure~\ref{fig:glass-door}, e.g., glass walls or uneven bumps on ground --- things which a robot can readily \textit{feel} as it walks through them.

\paragraphc{Proprioceptive Feedback:}
Our insight is to leverage this robot's on-ground \textit{feeling} as observed via \textit{proprioception} to bridge the gap and continually update the high-level navigation plan in accordance with low-level locomotion. Furthermore, this coupling of locomotion with navigation improves locomotion efficiency as well. For instance, a planner aware of locomotion ability can direct the robot to switch low-level gaits (walking $\rightarrow$ trotting $\rightarrow$ galloping) for increasing its speed whenever the path is straight and switch other way round to decrease speed on winding paths.  We posit that the adaptation of navigational plan from vision and proprioception must occur online in real time. But, how?

\paragraphc{Coupled Vision and Proprioception:}
We show a high-level illustration of our overall system \algo (Vision and Proprioception for Navigation) in Figure~\ref{fig:method}. It consists of three subsystems: a velocity-conditioned walking policy, a safety advisor module, and the planning module which together make synergistic use of vision and proprioception for navigation of legged robots.
At the lowest level, our velocity-conditioned locomotion controller is trained via reinforcement learning to allow the robot to walk at different speeds and in different directions. It takes the commanded linear and angular velocity as input along with the robot's proprioception state to predict the target joint angles directly without using any hand-engineered control primitives. We train this base controller in simulation via energy-based reward to allow for seamless gait switching at different speeds~\cite{fu2021minimizing,yang2021fast} and then transfer to the real world via rapid motor adaptation~\cite{rma} that estimates environment extrinsics using an adaptation module trained in simulation. 
Once we have learned the walking policy, which includes the base policy and the adaptation module, in simulation, we freeze it and train a Safety Advisor (SA) Module, also in simulation, which learns to estimate the safety constraints of the walking policy. It uses proprioception to estimate(1) if the robot is in collision to a visually undetected object such as glass walls (2) what is a safe velocity limit for the robot to walk in the current terrain which could be soft, slippery, bumpy, etc.
During deployment, the walking policy (base policy and adaptation module) and the SA (safety advisor) module are kept frozen and interact with the planner as shown in Figure~\ref{fig:method}. The planner uses on board cameras to compute a navigation cost map for an input to the point goal and takes in the two bits of safety constraints from the safety advisor module to compute the target linear and angular velocity which is given to the walking policy to track. This planner also ensures that both the linear and angular commanded velocities are within the feasible range of the walking policy.
The planning module continually updates the cost map and safety constraints to generate the target velocity for the walking velocity as the robot moves. All the modules run asynchronously onboard of the robot.

\paragraphc{Simulation and Real-World Evaluation:}
We evaluate our system \algo in challenging navigation settings (e.g., Figure~\ref{fig:teaserreal}) with difficult terrains, invisible glass obstacles, slippery surfaces, deformable ground and challenging outdoor scenarios. Please see videos at \url{\pagelink}

In addition, we conduct a series of experiments in simulation. For this we import real-world Matterport 3D \cite{Matterport3D} maps used in Habitat \cite{habitat19iccv} and Gibson \cite{xia2018gibson} into RaiSim to create a simulation benchmark for controlled study of joint navigation and legged locomotion. We find that the proposed system is 7\% - 15\% better than baselines with disjoint planning and control loop in different terrains and in settings with invisible obstacles. We find that minimizing time to goal can lead to more energy consuming behaviours which can be compensated for by the use of efficient locomotion policy with emergent gaits. We also additionally show the importance of legged systems over wheeled counterparts in traversing challenging terrains, and empirically demonstrate that continuous velocity-conditioned policy is more time efficient than its discrete counterpart. 

\vspace{-0.2em}
\section{Velocity-Conditioned Walking Policy}
\vspace{-0.2em}
Our velocity-conditioned walking policy is an implementation of the approach in ~\cite{rma,fu2021minimizing}. We present a review here to make this paper self-contained. The walking policy contains a base policy which takes the command velocity and the robot state as input and predicts the target joint angles. It additionally takes the extrinsics vector as input which is estimated by the adaptation module and enables rapid online adaptation to varying environment conditions ~\cite{rma}.

\paragraphc{Base Policy:}
We first train a base policy to walk in simulation on varying terrains and track a commanded linear and angular velocity. The base policy $\pi$ takes the current propriopceptive state (incl. joint angles, joint velocities, body row angle, body pitch angle, and foot contact indicators) $x_t \in \mathbb{R}^{30}$, command velocities $[v^{\mathrm{cmd}}, \omega^{\mathrm{cmd}}] \in \mathbb{R}^{2}$, previous action $a_{t-1} \in \mathbb{R}^{12}$, and the extrinsics vector $z_t \in \mathbb{R}^{8}$ to predict the target joint positions $a_t$, which are converted to torques by a PD controller. The extrinsics vector $z_t$ is an encoding of the environment conditions (like payload, friction, etc.) which enables the base policy to adapt to different environment conditions instead of being blind to it. The extrinsics vector $z_t$ is generated by an environment encoder $\mu$ from privileged environment information $e_t \in \mathbb{R}^{19}$,
as follows: $z_t = \mu(e_t)$ and $a_t = \pi(x_t, a_{t-1}, z_t)$.

We jointly train both $\pi$ and $\mu$ end-to-end with model-free reinforcement learning to maximize discounted expected return $J(\pi) = \mathbb{E}_{\tau \sim p(\cdot|\pi)}\left[\sum_{t=0}^{T-1}\gamma^t r_t\right]$, where $\tau = \{(x_0, a_0, r_0), . . ., (x_{T-1}, a_{T-1}, r_{T-1})\}$ is a sampled trajectory of the robot when executing policy $\pi$ in the simulation, and $p(\tau|\pi)$ represents the likelihood of the trajectory under $\pi$. We use PPO ~\cite{schulman2017proximal} to maximize this objective. 

\paragraphc{RL Reward:} Reward encourages the policy to accurately track a commanded linear and angular velocity while penalizing a higher energy consumption~\cite{fu2021minimizing}. We denote the linear velocity as $v$, the orientation as $\theta$ and the angular velocity as $\omega$, all in the robot's base frame. We additionally define the joint angles as $\bm{q}$, joint velocities as $\bm{\dot{q}}$, and joint torques as $\bm{\tau}$. The reward at time $r_t$ is defined as the sum of the following quantities (see supplementary for specifics):
\begin{itemize}[noitemsep,leftmargin=1.3em,itemsep=0em,topsep=.1em]
    \item Velocity Matching: $- | v_x  - v^{\mathrm{cmd}} | - |\omega_{\mathrm{yaw}} - \omega^{\mathrm{cmd}}|$
    \item Energy Consumption: $- {\bm{\tau}^T \bm{\dot{q}}}$
    \item Lateral Movement: $- |v_y| ^ 2$
    \item Hip Joints: $- \|\bm{q}_{\mathrm{hip}}\| ^ 2$
\end{itemize}
\vspace{-0.2em}

\paragraphc{Training Scheme:} Similar to ~\cite{rma}, we train our agent on fractal terrains without any additional artificial rewards for foot clearance or external pushes. For target velocities, we sample from one of the two settings: jointly track linear and angular velocity (curve following), or turning in place. Turning in place is important to handle very cluttered environments. See supplementary for range details.

\paragraphc{Adaptation Module:} 
Since we don't have the privileged environment information during deployment, we use RMA~\cite{rma} to train an adaptation module $\phi$ in simulation itself to estimate the extrinsics $z_t$ from proprioceptive state, which is available during deployment. Concretely, the adaptation module uses the recent history of robot's states $x_{t-k:t-1}$ and actions $a_{t-k:t-1}$ to generate $\hat{z_t}$ which is an estimate of the true extrinsics vector $z_t$. This is trained via supervised learning because we have access to both proprioceptive history and the true extrinsics vector in simulation. 

\vspace{-0.2em}
\section{Safety Advisor Module}
\vspace{-0.2em}
The safety advisor module captures the constraints which enable the robot to walk safely. For this, we train two safety advisors in simulation: (1) Collision Detector $M_c$ to detect collisions and (2) Fall Predictor $M_f$ to predict future falls, both from proprioception which includes the recent history of states ($x_{t-k:t-1}$) and actions ($a_{t-k:t-1}$) (analogous to ~\cite{rma} ). During deployment, the safety advisor module uses the prediction of these two advisors to inform the planner of the safe operating constraints of the walking policy. 

\paragraphc{Collision Detector ($M_c$):}
The collision detector estimates the probability of whether the robot is currently in collision, using proprioception ($M_c(x_{t-k:t-1}, a_{t-k:t-1})$). If a collision probability is above a threshold (0.5), the safety advisor module adds a fixed size patch of obstacle (9cm x 3cm, about the head size of A1), where the side with 3cm is in the current direction of robot, to the cost map in front of the current position of the robot to indicate an obstacle which may be missed by the vision system (e.g. glass walls). 

\paragraphc{Fall Predictor ($M_f$):}
The fall predictor makes a probability prediction of whether the walking policy is likely to fall within the next 1s using proprioception ($M_f(x_{t-k:t-1}, a_{t-k:t-1})$). If a fall probability is above a threshold (0.5), the safety advisor module decreases the velocity limit ($v^{max}_t$) by 0.2 m/s, otherwise it increases the velocity limit by 0.05 m/s. The planner uses $v^{max}_t$ to generate the linear velocity command for the walking policy. This enables the planner to slow the robot down in dangerous settings like soft or slippery terrains, heavy payload, etc.

\paragraphc{Module Training:}
We train both the safety advisors $M_f$ and $M_c$ in a self-supervised fashion in simulation. We collect data under randomly sampled environments and commands, and record the binary labels on (1) robot is currently in collision (2) if the policy results in a fall in the next 1s. We then train the safety advisors by minimizing binary cross-entropy loss. Details are in the supplementary.

\vspace{-0.2em}
\section{Visual Planner}
The visual planner uses the onboard cameras to generate a top down 2D cost map and uses it to plan a path to the goal. It additionally uses the safety constraints estimated by the safety advisor to generate the command velocities which are fed into the walking policy. Concretely, the visual planner consists of (1) a mapping module which generates a top down 2D occupancy map from onboard cameras, (2) cost map generation step using Fast Marching Method (FMM) and signed distance field, (3) PID based planner to use the cost map and safety constraints from the safety advisor module to generate linear and angular velocity commands for the walking policy.       

\subsection{Visual Occupancy Map}
We first generate a top down 2D visual occupancy map by incrementally accumulating point clouds from an onboard Intel RealSense D435 depth camera~\cite{keselman2017intel} as the robot moves. The point clouds are transformed into the world reference frame using pose information from an onboard tracking camera (Intel RealSense T265). The transformed point clouds are capped by a maximum height of interest and then dynamically projected into a horizontal 2D frame to form an occupancy map where each grid has a value from 0 to 1 to indicate the probability of being free space. The occupancy map is binarized for the path planning using a threshold of 0.5. We use an open-sourced implementation from Intel RealSense to compute the visual occupancy map~\cite{librealsense}. We convert it to a configuration space by modeling the robot size as a square and dilating the occupancy map.

\subsection{Cost Map Generation}
The 2D cost map is a sum of goal distance map (geodesic distance to the goal) and obstacle distance map (to maintain a safety margin from obstacles). Following the direction of steepest descent from any starting point in this cost map gives an obstacle free path to the goal.  

\paragraphc{Goal Distance Map:}
We use Fast Marching Method (FMM)~\cite{sethian1999fast} to compute the geodesic distance to the point goal, $d^{\mathrm{goal}}(x, y)$ for every starting position $(x, y)$. 

\paragraphc{Obstacle Distance Map:} We first compute the signed distance (L1 norm) from the closest obstacle for every point ($d^{\mathrm{sdf}}(x, y)$), and then compute the obstacle distance map as $\max (0, \alpha_1 - d^{\mathrm{sdf}}(x, y))$, where $\alpha_1$ is a distance threshold. We only penalize the robot when it is within $\alpha_1$ to an obstacle. This inverse signed distance field serves two purposes: 1) it penalizes the robot for being too close to obstacles; 2) gives a smooth differentiable cost map even at at (otherwise non-differentiable) object boundaries which enables smooth continuous path planning. 

\paragraphc{Cost Map:} The final cost map is
\vspace{-0.5em}
\begin{align}
    C(x, y) &= d^{\mathrm{goal}}(x, y) + \alpha_2 \max (0, \alpha_1 - d^{\mathrm{sdf}}(x, y))  \label{eq:fmm}
\end{align}

Here, $\alpha_2$ is a scaling factor to trade off the two costs. During deployment, the safety advisor module asynchronously adds an additional local obstacle to the cost map if the collision detector ($M_c$) predicts a collision.

\subsection{Velocity Command Generation}
Given the robot's current position $(x_t, y_t)$, heading (yaw) $\theta_t$ and cost map $C(x, y)$, the optimal heading direction is the direction of steepest descent in the cost map ~\cite{sethian1999fast}. We can compute this optimal heading direction or target orientation of the robot $\theta^{\mathrm{target}}_t$ as the normalized negative gradient of the cost map $-\nabla C(x_t, y_t)$.

\paragraphc{Angular Velocity:} 
We use a PD controller to compute the command angular velocity (\ref{fig:command_vel}) which is then clipped to the feasible range (specified in supplementary):
\vspace{-0.5em}
\begin{align}
\omega^{\mathrm{cmd}}_t = K_p \cdot (\theta^{\mathrm{target}}_t - \theta_t) + K_d \cdot (\omega^{\mathrm{target}}_t - \omega_t)
\end{align}

\vspace{-0.1in}
\paragraphc{Linear Velocity:}
We do a linear search in the cost map starting from the robot current position $(x_t, y_t)$ in the direction of $\theta$ to get a short-term target position $(x'_t, y'_t)$. The key insight is that the robot should go in its current direction as far as possible as long as the cost keeps decreasing (Figure~\ref{fig:command_vel}). The target linear velocity $v^{\mathrm{cmd}}$ is $\frac{1}{T}\alpha_0$, where $\alpha_0$ is obtained from the optimization problem in Figure \ref{fig:command_vel}a. A larger $T$ will lead to a more conservative target linear velocity, whereas a small $T$ will be more aggressive. The command sent to the robot is an exponentially smoothed average of the target speed. We maintain a separate exponential moving average for speed-up and slow-down. 

\begin{figure}[t]
    \vspace{-1em}
  \centering
  
    \includegraphics[width=1.0\linewidth]{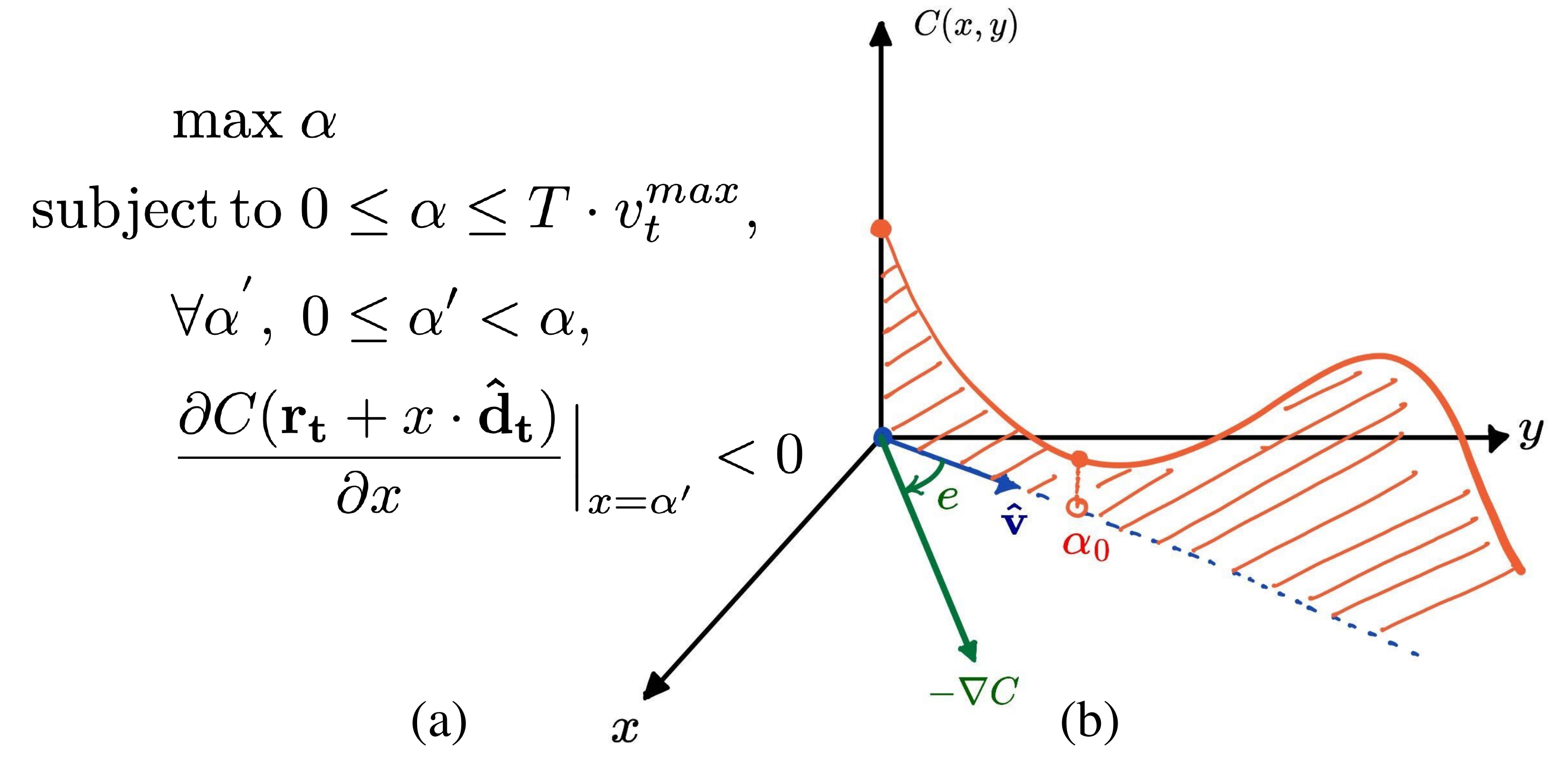}
\vspace{-2.2em}
  \caption{\small The optimal direction is along the direction of steepest descent in the cost map $-\nabla C$. The angular velocity is computed by PD control on the error $e$ between optimal direction and current direction $\mathbf{\hat{v}}$. The magnitude of linear velocity is determined by finding the furthest point $\alpha_0$ along the current direction such that the cost keeps decreasing. ($\mathbf{r_t}$: robots current position, $\mathbf{\hat{d_t}}$: unit vector in direction $\theta_t$, $v^{max}_t$: maximum linear walking speed from the fall predictor $M_f$, and $T$: lookahead time)}
  \label{fig:command_vel}
  \vspace{-1em}
 \end{figure}

\vspace{-0.2em}
\section{Experimental Setup}
\vspace{-0.2em}
\label{sec:setup}

\paragraphc{Physical Hardware:}
We use the A1 robot from Unitree with 18-DoF (12 actuatable). Its proprioception sensors include joint motor encoders, roll and pitch from the IMU sensor and  binarized foot contact indicators. We additionally mount Intel RealSense depth D435 and tracking T265 cameras. The deployed policy uses joint position control. 

\vspace{-0.04in}\paragraphc{Locomotion Policy:} 
For locomotion policy, we use similar architecture and training details as~\cite{rma, fu2021minimizing}, and list the exact policy and training details in the supplementary.

\vspace{-0.04in}\paragraphc{Safety Advisor Module:} 
Similar to the adaptation module, both the collision detector and fall predictor module share the same architecture and embed states and actions into a 32-dim vector using a linear layer. Then, we use 3 layers of 1D convolutions with input channels, output channels and strides $[32, 32, 8, 4], [32, 32, 5, 1], [32, 32, 5, 1]$. The flattened features are then passed through a 2-layer MLP with 8 hidden units to get 1 sigmoid output as the predicted probability value. We train the module in an online fashion by rollouts in environments with randomly sampled invisible obstacles, frictions, terrain roughness and payload values (see supplementary for ranges). At simulation test time, we run both the collision detector and fall predictor at 5Hz, whereas for deployment on robot we train a lightweight version using only the last $0.2$s of observation history and run it at 10Hz. More details are in the supplementary. 

\vspace{-0.04in}\paragraphc{FMM Planner and PID Controller:} 
During cost map generation we choose $\alpha_1 = 0.3m, \alpha_2 = 0.5$. For controlling the angular velocity we set gains $K_p = 1, K_d = 0.02$ with $\omega^{\mathrm{target}}$ set to 0. At run time, we clip the linear speed supplied by the line search to the maximum command velocity determined by the fall predictor. To facilitate in-place turning, if the linear speed is less than $0.2$, we clip the angular velocity to the range $[0.4, 0.8]$. The planner runs at 10Hz in simulation and robot. 

\begin{figure}[t]
\vspace{-1em}
\centering
\includegraphics[width=\linewidth]{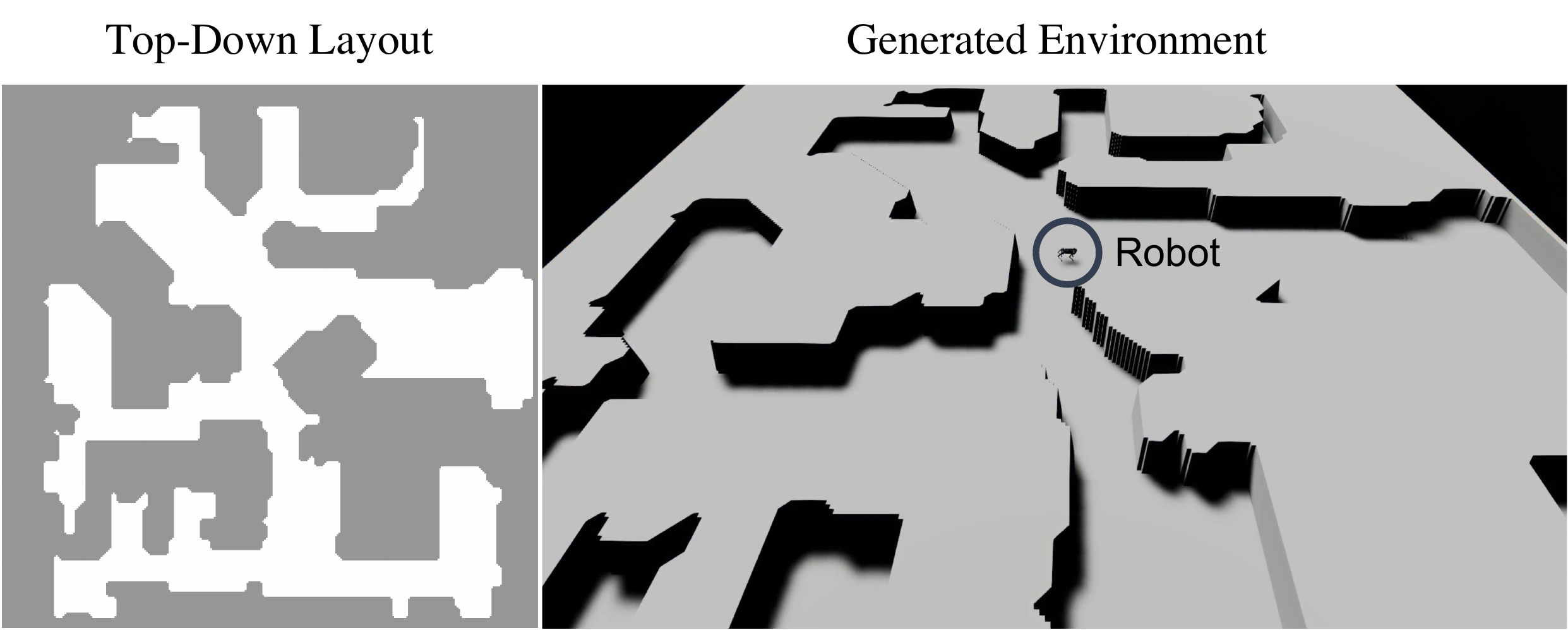}
\vspace{-2em}
\caption{\small An example of the top-down view room layout and the corresponding generated simulation environment.}
\vspace{-1.4em}
\label{fig:sim_layout}
\end{figure}

\vspace{-0.04in}
\paragraphc{Simulation Environments:} 
We generate top-down view room layouts from room scanning meshes using habitat-sim~\cite{szot2021habitat,habitat19iccv}.
The meshes are from gibson environment~\cite{xia2018gibson} and matterport3D~\cite{Matterport3D}. We then select 200 challenging room layouts for navigation as our validation set. For each room layout, we sample 10 navigation goals and set the initial point to be the farthest point from the goal. We then convert the room layout to RaiSim simulation environment~\cite{hwangbo2018per}. The resolution is 0.1m per pixel. We show an example of the top-down layouts and the generated environment in figure~\ref{fig:sim_layout}.

To demonstrate our navigation system on complex terrains, we construct the following variations:
\begin{itemize}[noitemsep,leftmargin=1.3em,itemsep=0em,topsep=.1em]
    \item Flat: flat surface with coefficient of friction $\mu=0.8$.
    \item RoughTerrain: we put eight patches of z-scale 0.05 and size 0.8m$\times$0.8m along the path from initial to the goal position. The rough terrain is constructed using the built-in terrain generator by RaiSim~\cite{hwangbo2018per}.
    \item 2x/4x/8x Inv-Obstacle: we put 2/4/8 0.2m$\times$0.2m obstacles that cannot be detected by the vision sensor.
    \item Randomized: we put 8 rough and slippery patches along the path from initial to goal position. The rough patches are of z-scale 0.05. The coefficient of friction of slippery patches are sampled from \{0.1, 0.3, 0.5, 0.7, 0.9\}. An 8kg payload (A1 itself is ~12kg) is placed on / removed from top of the robot every 5s. 
\end{itemize}

\vspace{-0.2em}
\section{Experimental Results}
\noindent We test our approach both in simulation and in the real world. 

\begin{figure*}[t]
    \vspace{-0.4cm}
    \centering
\includegraphics[width=1.0\linewidth]{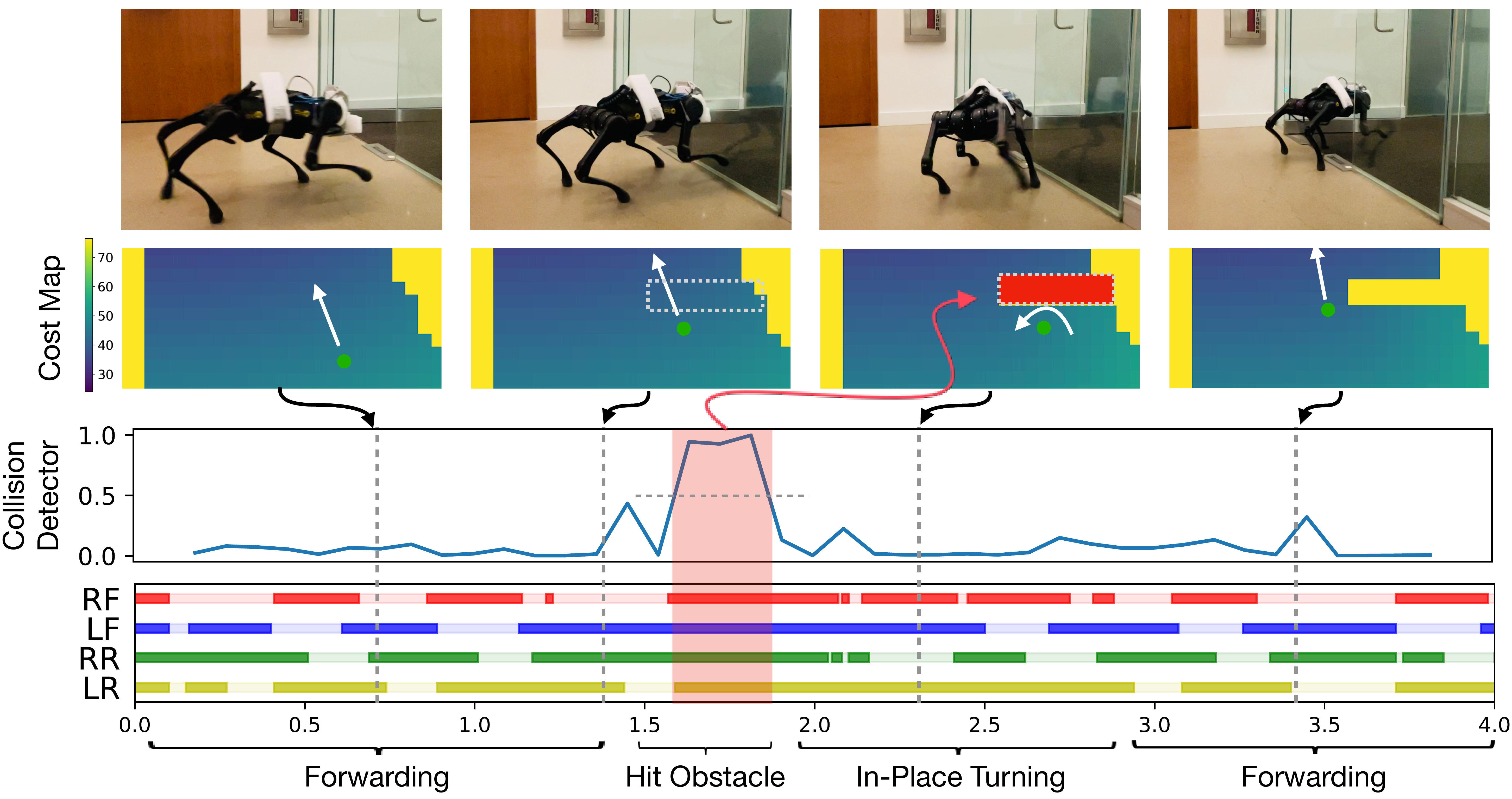}
    \vspace{-2.0em}
    \caption{\small Collision Detector: The top row shows the deployed robot, the second row shows the state of the occupancy map and the bottom two rows show the predictions of the collision detector and the gait plot of the robot. The robot collides with the glass wall which is missed by the onboard cameras, after which, the collision detector detects this from proprioception and indicates a missed obstacle. The map is updated locally to indicate this and the robot replans its path around it. The gait plot shows that the robot is stuck for a fraction of the second before the collision detector senses the glass wall and updates the map. \algo bypasses the glass wall with a 100\% (8 out of 8) success rate, whereas a vision only baseline fails to cross it even once.}
    \label{fig:glass-door}
    \vspace{-0.5cm}
\end{figure*}

\subsection{Simulation Experiments}
In simulation we assume that the agent has access to the ground-truth occupancy map, and we only vary the terrains and the navigation strategy. The purpose of our simulation experiments is to answer the following questions:
\begin{itemize}[noitemsep,leftmargin=1.3em,itemsep=0em,topsep=.1em]
    \item How much does proprioception feedback help?
    \item Minimizing time to goal requires more aggressive walking and more energy. Can a varying gait policy compensate for some of the energy consumed?
\end{itemize}
We additionally evaluate the following broader questions:
\begin{itemize}[noitemsep,leftmargin=1.3em,itemsep=0em,topsep=.1em]
\item Does legged locomotion improve goal reaching?
\item Is continuous velocity conditioning better than discrete?  
\end{itemize}

\paragraphc{Baseline and Metrics:} We use the LoCoBot~\cite{locobot}
as our wheeled robot baseline, as it is widely used in visual navigation~\cite{chaplot2020learning,chaplot2020object,bansal2020combining,gupta2017cognitive}. We import the PyRobot URDF model~\cite{pyrobot2019}. Both \algo and the LoCoBot use a control frequency of 100Hz and a planning frequency of 10Hz. We evaluate our system using the following metrics: 1) Success Rate 2) Success weighted by (normalized inverse) Path Length (SPL)~\cite{anderson2018evaluation} 3) Average time used to achieve the goal. If the agent fails to reach the goal, we add a constant timeout penalty (220s) for the failure episodes; 4) Average energy consumption over the successful episodes~\cite{fu2021minimizing}.

\paragraphc{Improvements with Proprioceptive Coupling:} We separately analyze the importance of the two safety advisors (collision detector and fall predictor). 

\textit{Collision Detector:} We uniformly place 2/4/8 0.2m$\times$0.2m obstacles along the path from initial and goal positions, and run \algo with/without proprioceptive feedback. The obstacles are not marked in the top-down view map to simulate the glass or other objects that an imperfect vision sensor fails to capture. In Table~\ref{tab:invisible_obs}, we note that adding invisible obstacles makes the navigation task very challenging as evident from the performance drop of all the methods. Using the proprioceptive collision detector module improves the success rate by 5.7 points over the baseline method which does not use it. The performance improvements are even larger when the environment becomes more challenging with up to 15 points improvement over baseline.

\textit{Fall Predictor:} In Table~\ref{tab:slippery}, we show that learned fall predictor enables safe navigation in challenging environments involving a combination of slippery surfaces, rough terrains, and payload changes. We put eight 2.4m$\times$2.4m patches with uneven slippery surfaces along the path from initial and the goal positions. An 8kg payload is placed / removed to the robot every 5 second. Using the proprioceptive fall prediction to adjust the speed of the robot gives 7 points higher goal-reaching success rate over the baseline without proprioception.

\paragraphc{Compensating for higher energy consumption induced by minimizing time to goal:} Minimizing time to goal leads to aggressive locomotion behaviours and increased energy consumption. To compensate for some of the increase in energy consumption, we show that a policy with efficient gaits ~\cite{fu2021minimizing} leads to a 10\% lower energy consumption compared to a fixed gait trotting-only policy (Table~\ref{tab:energy}). \algo also has a slightly higher success rate because it switches to a more stable gait at low speeds when traversing complex settings, as compared to a fixed gait policy. \algo automatically switches gaits to optimize for stability and energy at different speeds. 

\begin{table}[t]
    \centering
    \resizebox{\linewidth}{!}{
    \setlength{\tabcolsep}{4pt}
    \renewcommand{\arraystretch}{1.3}
    \begin{tabular}{c|c|c|ccc}
    & Navigation System & Terrain Type & Success $\uparrow$ & SPL $\uparrow$  & Time(s) $\downarrow$  \\ %
    \hline
    \hline
    (a) & w/o Proprio & Flat & 95.20 & 0.79 & 80.28 \\ %
    \hline
    (b) & w/o Proprio & 2x Inv-Obstacle & 68.45 & 0.57 & 119.80 \\
    (c) & \algo (Ours) & 2x Inv-Obstacle & \textbf{74.15} & \textbf{0.61} & \textbf{111.93} \\ %
    \hline
    (d) & w/o Proprio & 4x Inv-Obstacle & 45.85 & 0.38 & 152.39 \\ %
    (e) & \algo (Ours) & 4x Inv-Obstacle & \textbf{59.20} & \textbf{0.49} & \textbf{134.70} \\ %
    \hline
    (f) & w/o Proprio & 8x Inv-Obstacle & 24.35 & 0.20 & 184.07 \\ %
    (g) & \algo (Ours) & 8x Inv-Obstacle & \textbf{39.25} & \textbf{0.32} & \textbf{164.95} \\ %
    \end{tabular}
    }
    \vspace{-1em}
    \caption{\small \textbf{Proprioceptive feedback helps navigation with invisible obstacles.} With proprioceptive feedback, the Success Rate is improved by more than 5 points when two invisible obstacles present. In more challenging environment, the performance improvement is increased to 15 points.}
    \vspace{-1em}
    \label{tab:invisible_obs}
\end{table}

\begin{table}[t]
    \centering
    \resizebox{\linewidth}{!}{
    \setlength{\tabcolsep}{4pt}
    \renewcommand{\arraystretch}{1.3}
    \begin{tabular}{c|c|c|ccccc}
    & Navigation System & Terrain Type & Success $\uparrow$ & SPL $\uparrow$   & Time(s) $\downarrow$  \\
    \hline
    \hline
    (a) & w/o Proprio & Flat & 95.20 & 0.79 & 80.28 \\ %
    \hline
    (b) & w/o Proprio & Randomized & 80.25 & 0.66 & \textbf{105.68} \\ %
    (c) & \algo (Ours) & Randomized & \textbf{87.40} & \textbf{0.73} & 117.65 \\
    \end{tabular}
    }
    \vspace{-1em}
    \caption{\small \textbf{Proprioceptive feedback helps navigation with challenging terrains.} Without proprioceptive feedback, the success rate is decreased by 7 points in the presence of a combination of slippery, rough surfaces, and abrupt payload changes, which cannot be inferred from a vision-only system. But with proprioception, the planner can readily ``feel'' the terrain property and the payload changes and plan with a safer velocity.}
    \vspace{-1.0em}
    \label{tab:slippery}
\end{table}

\paragraphc{Legs vs. Wheels:} We also compare \algo with LoCoBot on visual navigation in Table~\ref{tab:loco_rough}. 
On flat terrains, LoCoBot has a slightly lower performance since LoCoBot is more prone to getting stuck in the local minima of the FMM map (see supplementary for details).
Whereas adding rough terrain (5cm elevation) to the environment leads to a significant drop in goal-reaching performance of the LoCoBot. We additionally try the planning scheme which plans around rough terrains while assuming ground truth access to their locations. Although the success rate improves, the time cost is still significantly worse than our legged robot baseline, which is able to maintain similar success rate and time to goal because of its robust walking capabilities. In short, though energy efficient, wheeled robots struggle on uneven terrains, whereas legged robots are more terrain-agnostic. 

\paragraphc{Continuous Velocity Conditioning vs. Discrete:} We compare our continuous planner to a discrete planner typically used in visual navigation~\cite{gupta2017cognitive, savinov2018semi, mirowski2018learning, kumar2018visual}. Our discrete planner only commands four actions: 1) forward with 0.6 m/s; 2) turn left with 0.8 rad/s; 3) turn right with 0.8 rad/s; 4) stop, whereas planning over the continuous range of linear and angular velocities enables smoother trajectory and shorter time to goal. In Table~\ref{tab:planner}, we see that our system \algo is $27\%$ more time efficient than a discrete planner as the robot can simultaneously turn and go forward.

\subsection{Real-World Experiments}
\vspace{-0.1cm}

\begin{figure*}[t]
    \vspace{-0.7cm}
    \centering
\includegraphics[width=1.0\linewidth]{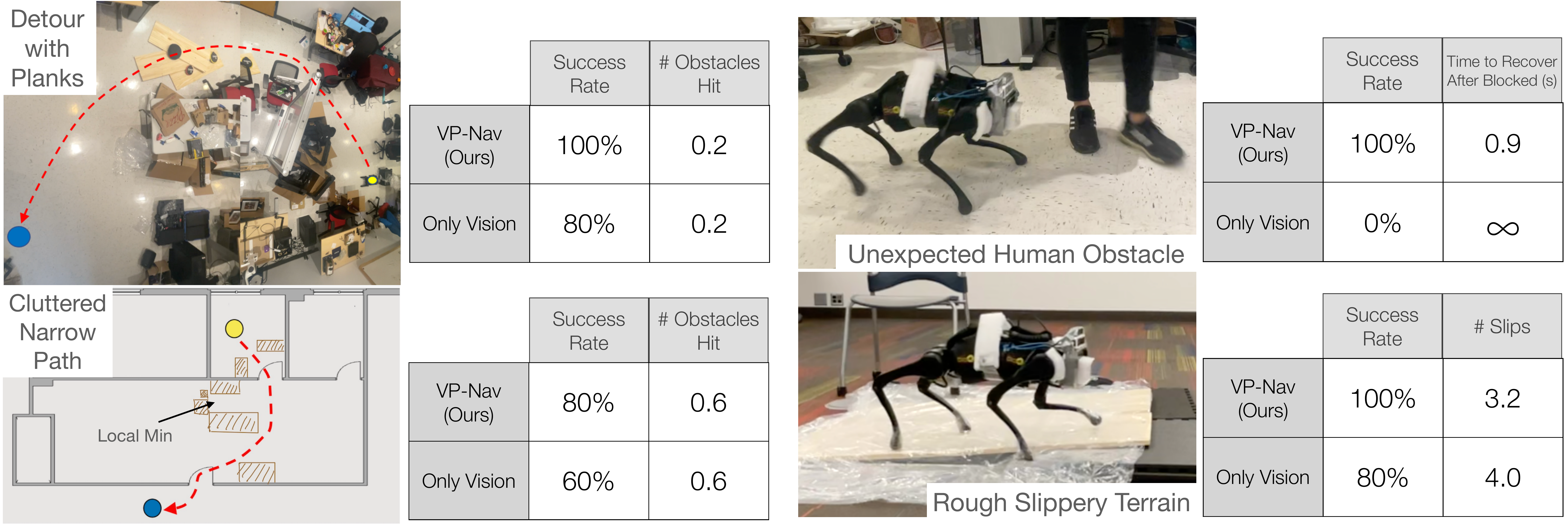}
    \vspace{-0.65cm}
    \caption{\small Real-World Experiments: We compare \algo with a pure vision approach (without proprioceptive feedback from safety advisor) and evaluate for 5 trials in several challenging settings. \algo gives a higher success rate in all these settings. On the left, we have 2 indoor tasks with planks scattered on the floor and cluttered narrow paths. In both settings, texture-less walls, transparent panels, large brown packaging boxes can be missed by vision. With proprioceptive feedback from safety advisor, we update the occupancy map and replan, despite hitting same number of obstacles. On the right, we tested with a fast human obstruction. With proprioceptive feedback from safety advisor, the robot recovers within a second. Additionally, on challenging terrains as shown on the bottom right, a likely fall detected by the predictor can be used to decrease the safe velocity limit, and improve the stability and success rate.}
    \vspace{-0.4cm}
    \label{fig:real-world-table}
\end{figure*}

\paragraphc{Invisible Obstacles: } We tested the collision detector with invisible obstacles like glass doors, humans that abruptly walk into the robot's path, walls and boxes without textures (Fig~\ref{fig:glass-door}, ~\ref{fig:real-world-table}). We find that feedback from the safety module gives higher success rate in all these settings. The glass wall, which is invisible to the onboard cameras is detected by the proprioceptive feedback once the robot collides with the door. The missed obstacle is then updated in the map at the place of the collision and the robot replans its path around it. Humans abruptly rushing into the robot's path are similarly missed by the camera, and then later block the robot's cameras to be detected by them (depth camera's near distance is around 30cm). Such obstacles that suddenly appear from outside into the field-of-view render the trajectory prediction approaches useless~\cite{hoeller2021learning}. With proprioceptive collision detector, our robot can reason about these ``invisible" objects and update its occupancy map to plan a new path.

\begin{table}[t]
    \centering
    \resizebox{\linewidth}{!}{
    \setlength{\tabcolsep}{4pt}
    \renewcommand{\arraystretch}{1.3}
    \begin{tabular}{c|c|c|ccc}
    & Navigation System & Terrain Type & Success $\uparrow$ & SPL $\uparrow$   & Energy(K) $\downarrow$  \\
    \hline
    \hline
    (a) & Trot Only & Flat & 93.80 & 0.77 & 252.56 \\
    (b) & \algo (Ours) & Flat & \textbf{95.20} & \textbf{0.79} & \textbf{233.05} \\
    \end{tabular}
    }
    \vspace{-1em}
    \caption{\small \textbf{Energy efficiency}. Our policy with varying gaits consumes less energy compared with the single-gait policy.}
    \label{tab:energy}
    \vspace{-1.2em}
\end{table}

\begin{table}[t]
    \centering
    \resizebox{\linewidth}{!}{
    \setlength{\tabcolsep}{4pt}
    \renewcommand{\arraystretch}{1.3}
    \begin{tabular}{c|c|c|ccc}
    & Navigation System & Terrain Type & Success $\uparrow$ & SPL $\uparrow$   & Time(s) $\downarrow$  \\
    \hline
    \hline
    (a) & LoCoBot-Proceed & Flat & 90.65 & \textbf{0.81} & 102.98 \\ %
    (b) & \algo (Ours) & Flat & \textbf{95.20} & 0.79 & \textbf{80.28} \\ %
    \hline
    (c) & LoCoBot-Proceed & RoughTerrain & 15.70 & 0.14 & 215.28 \\ %
    (d) & LoCoBot-Avoid & RoughTerrain & 69.10 & 0.60 & 146.85 \\ %
    (e) & \algo (Ours) & RoughTerrain & \textbf{95.05} & \textbf{0.79} & \textbf{80.87} \\ %
    \end{tabular}
    }
    \vspace{-1em}
    \caption{\small \textbf{The importance of legs for goal-reaching.} LoCoBot cannot easily pass rough terrains even when its height is only 5cm. The success rate drops to only 15.7 (LoCoBot-Proceed). Even when the LoCoBot has access to location of rough terrain patches and can plan to avoid it (LoCoBot-Avoid), the success rate is still significantly lower than ours with a higher time cost.}
    \vspace{-1em}
    \label{tab:loco_rough}
\end{table}

\begin{table}[t]
    \centering
    \small
    \setlength{\tabcolsep}{4pt}
    \renewcommand{\arraystretch}{1.3}
    \resizebox{\linewidth}{!}{
    \begin{tabular}{c|c|c|ccc}
        & Navigation System & Terrain Type & Success $\uparrow$ & SPL $\uparrow$  & Time(s) $\downarrow$  \\ %
        \hline
        \hline
        (a) & LoCoBot-Dis & Flat & 86.45 & 0.77 & 178.27 \\ %
        (b) & LoCoBot-Cts & Flat & \textbf{90.65} & \textbf{0.81} & \textbf{102.98} \\ %
        \hline
        (c) & \algo -Dis & Flat & 95.35 & 0.80 & 110.27 \\ %
        (d) & \algo -Cts (Ours) & Flat & 95.20 & 0.79 & \textbf{80.87} \\ %
        \end{tabular}
    }
    \vspace{-1em}
    \caption{\small \textbf{Comparison between discrete planner (-Dis) and continuous planner (-Cts)}. Using the continuous planner makes the navigation system spend less time to reach the goal.
    }
    \vspace{-1.5em}
    \label{tab:planner}
\end{table}

\paragraphc{Rough Slippery Terrains: } We tested the fall detector with challenging terrains including movable planks scattered on the floor and slippery terrain, shown in Figure~\ref{fig:real-world-table} and in the supplementary. On rough slippery terrain, the fall predictor uses proprioception to estimate the risk of falling and accordingly decreases the velocity to ensure safety.  

\paragraphc{Other Complex Indoor Navigation:} We deploy \algo in challenging settings and compare to baselines which use pure vision without fall prediction and collision detection from proprioceptive feedback, and evaluate for 5 trials in all settings (Figure~\ref{fig:real-world-table}). We find that using vision and proprioception for coupled navigation and locomotion gives a higher success rate in all these settings. In the left of Figure~\ref{fig:real-world-table}, we have 2 indoor tasks which require taking a detour with planks scattered on the floor and maneuvers through a cluttered narrow path. In both settings, there are objects that can easily be missed by the vision system, including white walls with no texture, transparent desktop side panels and large brown packaging boxes in dim light. With the proprioceptive safety advisor, our robot can reason about these ``invisible" objects and update its occupancy map to replan for a new viable path, despite hitting the same number of obstacles. The robot also slows down on unstable planks that are scattered on the ground.

\vspace{-0.5em}
\section{Related Work}
\vspace{-0.5em}
\label{sec:related}
\paragraphc{Visual Navigation:} 
Visual navigation is mainly studied on wheeled robots by chaining mapping, localizing, and planning.
Once a 2D map is created, an optimal path to a goal can be found using graph search techniques \cite{hart1968formal, stentz1997optimal, koenig2005fast, lavalle1998rapidly}, level-set methods \cite{kimmel1996fast} or potential field methods \cite{khatib1986potential, khatib1986real} among others. The map is constructed via simultaneous localization and mapping using classical \cite{thrun2002probabilistic, montemerlo2002fastslam, fuentes2015visual} or learned methods \cite{chaplot2020learning, parisotto2017neural, zhang2017neural, mezghani2021memory, zhao2021surprising, datta2020integrating, wijmans2019dd, bansal2020combining, zhu2017target, gupta2017cognitive, khan2017memory, chiang2019learning} assuming access to nearly perfect low-level control. In our benchmark, we import maps from the common navigation datasets include Habitat \cite{habitat19iccv}, Gibson \cite{xia2018gibson} and Matterport3D \cite{Matterport3D}.
\paragraphc{Navigation of Legged Robots:}
Earlier works decouple locomotion and navigation which restricts the application only to simple terrains \cite{wooden2010autonomous}. This decoupled framework has been extended to include learned modules for cluttered environment navigation \cite{truong2020learning, li2021planning, hoeller2021learning}. 
\cite{chestnutt2007navigation} describe a coupled navigation and locomotion framework by estimating foothold placements from an elevation map. Foothold scores can be estimated heuristically \cite{wermelinger2016navigation, chilian2009stereo, jenelten2020perceptive, mastalli2015line, fankhauser2018robust, kim2020vision} or learned \cite{kolter2008control, kalakrishnan2009learning, wellhausen2021rough, mastalli2017trajectory, magana2019fast}. Other methods forgo explicit foothold optimization and learn traversibility maps \cite{yang2021real, chavez2018learning, guzzi2020path}. Several works complement vision-based state estimation by using contact information \cite{hartley2018hybrid,hartley2020contact,hereid20163d,lew2019contact}. 
Instead of relying only on vision, we combine navigation and locomotion via coupling vision and proprioception.

\paragraphc{Legged Locomotion:} This has conventionally been accomplished using control theory \cite{miura1984dynamic,raibert1984hopping,geyer2003positive,yin2007simbicon,sreenath2011compliant,johnson2012tail,khoramshahi2013piecewise,ames2014rapidly,hyun2016implementation,barragan2018minirhex, bledt2018cheetah, hutter2016anymal, imai2021vision} over handcrafted dynamics models. 
Recently, RL has been successfully used to learn such policies in simulation ~\cite{schulman2017proximal,lillicrap2016continuous,mnih2016asynchronous,fujimoto2018addressing} and in the real world with sim2real methods \cite{tan2018sim,tobin2017domain,peng2018sim,xie2020dynamics,nachum2020multi, hwangbo2019learning,tan2018sim,hanna2017grounded}. Alternatively, a policy learned in simulation can be adapted at test-time to work well in real environments \cite{yu2017preparing, yu2018policy, peng2020learning, zhou2019environment, yu2019biped, yu2020learning, song2020rapidly, clavera2018learning, rma, fu2021minimizing, smith2021legged, yang2022learning}.

\vspace{-0.5em}
\section{Conclusion and Limitations}
\vspace{-0.5em}
The use of a legged robot instead of a wheeled one broadens the applicability of visual navigation to complex terrains and environments. In this paper, we combine low-level locomotion with high-level navigation planning to enable goal-reaching for a legged quadruped robot. Our approach, \algo, tightly couples vision and proprioception to exploit their complementary strengths for robust navigation in the presence of disturbances, transparent obstacles and complex terrains which may not be detected by vision alone. \algo is lightweight and only uses the modest onboard computation and storage of the low-cost A1 quadruped robot. One limitation of our system is that low-level locomotion module communicates with navigation planner via safety module, and is not conditioned on the vision directly. Due to this, the robot can walk around obstacles but can not climb or jump over them. We leave vision-guided locomotion for future.

\paragraphc{Acknowledgement} We thank Aravind Sivakumar, Kenny Shaw and Shivam Duggal for help in real-world experiments. This work is supported by DARPA Machine Common Sense program and in part by Good AI research award.

\label{sec:conclusion}

{\small
\bibliographystyle{ieee_fullname}
\bibliography{main}

\begin{thebibliography}{10}\itemsep=-1pt

\bibitem{locobot}
Locobot.
\newblock \url{http://www.locobot.org/}.

\bibitem{ames2014rapidly}
Aaron~D Ames, Kevin Galloway, Koushil Sreenath, and Jessy~W Grizzle.
\newblock Rapidly exponentially stabilizing control lyapunov functions and
  hybrid zero dynamics.
\newblock {\em IEEE Transactions on Automatic Control}, 2014.

\bibitem{anderson2018evaluation}
Peter Anderson, Angel Chang, Devendra~Singh Chaplot, Alexey Dosovitskiy,
  Saurabh Gupta, Vladlen Koltun, Jana Kosecka, Jitendra Malik, Roozbeh
  Mottaghi, Manolis Savva, and Amir~R. Zamir.
\newblock On evaluation of embodied navigation agents.
\newblock {\em arXiv:1807.06757}, 2018.

\bibitem{bansal2020combining}
Somil Bansal, Varun Tolani, Saurabh Gupta, Jitendra Malik, and Claire Tomlin.
\newblock Combining optimal control and learning for visual navigation in novel
  environments.
\newblock In {\em CoRL}, 2019.

\bibitem{barragan2018minirhex}
Monica Barragan, Nikolai Flowers, and Aaron~M. Johnson.
\newblock {MiniRHex}: A small, open-source, fully programmable walking hexapod.
\newblock In {\em RSS Workshop}, 2018.

\bibitem{bledt2018cheetah}
Gerardo Bledt, Matthew~J. Powell, Benjamin Katz, Jared Di~Carlo, Patrick~M
  Wensing, and Sangbae Kim.
\newblock Mit cheetah 3: Design and control of a robust, dynamic quadruped
  robot.
\newblock In {\em IROS}, 2018.

\bibitem{Matterport3D}
Angel Chang, Angela Dai, Thomas Funkhouser, Maciej Halber, Matthias Niessner,
  Manolis Savva, Shuran Song, Andy Zeng, and Yinda Zhang.
\newblock Matterport3d: Learning from rgb-d data in indoor environments.
\newblock In {\em 3DV}, 2017.

\bibitem{chaplot2020learning}
Devendra~Singh Chaplot, Dhiraj Gandhi, Saurabh Gupta, Abhinav Gupta, and Ruslan
  Salakhutdinov.
\newblock Learning to explore using active neural slam.
\newblock In {\em ICLR}, 2020.

\bibitem{chaplot2020object}
Devendra~Singh Chaplot, Dhiraj~Prakashchand Gandhi, Abhinav Gupta, and Ruslan
  Salakhutdinov.
\newblock Object goal navigation using goal-oriented semantic exploration.
\newblock In {\em NeurIPS}, 2020.

\bibitem{chavez2018learning}
R~Omar Chavez-Garcia, J{\'e}r{\^o}me Guzzi, Luca~M Gambardella, and Alessandro
  Giusti.
\newblock Learning ground traversability from simulations.
\newblock {\em RA-L}, 2018.

\bibitem{chestnutt2007navigation}
Joel Chestnutt.
\newblock {\em Navigation planning for legged robots}.
\newblock Carnegie Mellon University, 2007.

\bibitem{chiang2019learning}
Hao-Tien~Lewis Chiang, Aleksandra Faust, Marek Fiser, and Anthony Francis.
\newblock Learning navigation behaviors end-to-end with autorl.
\newblock {\em RA-L}, 2019.

\bibitem{chilian2009stereo}
Annett Chilian and Heiko Hirschm{\"u}ller.
\newblock Stereo camera based navigation of mobile robots on rough terrain.
\newblock In {\em IROS}, 2009.

\bibitem{clavera2018learning}
Ignasi Clavera, Anusha Nagabandi, Simin Liu, Ronald~S. Fearing, Pieter Abbeel,
  Sergey Levine, and Chelsea Finn.
\newblock Learning to adapt in dynamic, real-world environments through
  meta-reinforcement learning.
\newblock In {\em ICLR}, 2019.

\bibitem{datta2020integrating}
Samyak Datta, Oleksandr Maksymets, Judy Hoffman, Stefan Lee, Dhruv Batra, and
  Devi Parikh.
\newblock Integrating egocentric localization for more realistic point-goal
  navigation agents.
\newblock {\em arXiv:2009.03231}, 2020.

\bibitem{dudek2010computational}
Gregory Dudek and Michael Jenkin.
\newblock {\em Computational principles of mobile robotics}.
\newblock Cambridge University Press, 2010.

\bibitem{fankhauser2018robust}
P{\'e}ter Fankhauser, Marko Bjelonic, C~Dario Bellicoso, Takahiro Miki, and
  Marco Hutter.
\newblock Robust rough-terrain locomotion with a quadrupedal robot.
\newblock In {\em ICRA}, 2018.

\bibitem{fu2021minimizing}
Zipeng Fu, Ashish Kumar, Jitendra Malik, and Deepak Pathak.
\newblock Minimizing energy consumption leads to the emergence of gaits in
  legged robots.
\newblock In {\em CoRL}, 2021.

\bibitem{fuentes2015visual}
Jorge Fuentes-Pacheco, Jos{\'e} Ruiz-Ascencio, and Juan~Manuel
  Rend{\'o}n-Mancha.
\newblock Visual simultaneous localization and mapping: a survey.
\newblock {\em Artificial intelligence review}, 2015.

\bibitem{fujimoto2018addressing}
Scott Fujimoto, Herke Hoof, and David Meger.
\newblock Addressing function approximation error in actor-critic methods.
\newblock In {\em ICML}, 2018.

\bibitem{geyer2003positive}
Hartmut Geyer, Andre Seyfarth, and Reinhard Blickhan.
\newblock Positive force feedback in bouncing gaits?
\newblock {\em Proceedings of the Royal Society of London. Series B: Biological
  Sciences}, 2003.

\bibitem{gupta2017cognitive}
Saurabh Gupta, James Davidson, Sergey Levine, Rahul Sukthankar, and Jitendra
  Malik.
\newblock Cognitive mapping and planning for visual navigation.
\newblock In {\em CVPR}, 2017.

\bibitem{guzzi2020path}
J{\'e}r{\^o}me Guzzi, R~Omar Chavez-Garcia, Mirko Nava, Luca~Maria Gambardella,
  and Alessandro Giusti.
\newblock Path planning with local motion estimations.
\newblock {\em RA-L}, 2020.

\bibitem{hanna2017grounded}
Josiah Hanna and Peter Stone.
\newblock Grounded action transformation for robot learning in simulation.
\newblock In {\em AAAI}, 2017.

\bibitem{hart1968formal}
Peter~E Hart, Nils~J Nilsson, and Bertram Raphael.
\newblock A formal basis for the heuristic determination of minimum cost paths.
\newblock {\em IEEE transactions on Systems Science and Cybernetics}, 1968.

\bibitem{hartley2020contact}
Ross Hartley, Maani Ghaffari, Ryan~M Eustice, and Jessy~W Grizzle.
\newblock Contact-aided invariant extended kalman filtering for robot state
  estimation.
\newblock {\em The International Journal of Robotics Research}, 2020.

\bibitem{hartley2018hybrid}
Ross Hartley, Maani~Ghaffari Jadidi, Lu Gan, Jiunn-Kai Huang, Jessy~W Grizzle,
  and Ryan~M Eustice.
\newblock Hybrid contact preintegration for visual-inertial-contact state
  estimation using factor graphs.
\newblock In {\em IROS}, 2018.

\bibitem{hereid20163d}
Ayonga Hereid, Eric~A Cousineau, Christian~M Hubicki, and Aaron~D Ames.
\newblock 3d dynamic walking with underactuated humanoid robots: A direct
  collocation framework for optimizing hybrid zero dynamics.
\newblock In {\em ICRA}, 2016.

\bibitem{hoeller2021learning}
David Hoeller, Lorenz Wellhausen, Farbod Farshidian, and Marco Hutter.
\newblock Learning a state representation and navigation in cluttered and
  dynamic environments.
\newblock {\em RA-L}, 2021.

\bibitem{hutter2016anymal}
Marco Hutter, Christian Gehring, Dominic Jud, Andreas Lauber, C~Dario
  Bellicoso, Vassilios Tsounis, Jemin Hwangbo, Karen Bodie, Peter Fankhauser,
  Michael Bloesch, et~al.
\newblock Anymal-a highly mobile and dynamic quadrupedal robot.
\newblock In {\em IROS}, 2016.

\bibitem{hwangbo2019learning}
Jemin Hwangbo, Joonho Lee, Alexey Dosovitskiy, Dario Bellicoso, Vassilios
  Tsounis, Vladlen Koltun, and Marco Hutter.
\newblock Learning agile and dynamic motor skills for legged robots.
\newblock {\em Science Robotics}, 2019.

\bibitem{hwangbo2018per}
Jemin Hwangbo, Joonho Lee, and Marco Hutter.
\newblock Per-contact iteration method for solving contact dynamics.
\newblock {\em RA-L}, 2018.

\bibitem{hyun2016implementation}
Dong~Jin Hyun, Jongwoo Lee, SangIn Park, and Sangbae Kim.
\newblock Implementation of trot-to-gallop transition and subsequent gallop on
  the mit cheetah i.
\newblock {\em IJRR}, 2016.

\bibitem{imai2021vision}
Chieko~Sarah Imai, Minghao Zhang, Yuchen Zhang, Marcin Kierebinski, Ruihan
  Yang, Yuzhe Qin, and Xiaolong Wang.
\newblock Vision-guided quadrupedal locomotion in the wild with multi-modal
  delay randomization.
\newblock {\em arXiv:2109.14549}, 2021.

\bibitem{jenelten2020perceptive}
Fabian Jenelten, Takahiro Miki, Aravind~E Vijayan, Marko Bjelonic, and Marco
  Hutter.
\newblock Perceptive locomotion in rough terrain--online foothold optimization.
\newblock {\em RA-L}, 2020.

\bibitem{johnson2012tail}
Aaron~M Johnson, Thomas Libby, Evan Chang-Siu, Masayoshi Tomizuka, Robert~J
  Full, and Daniel~E Koditschek.
\newblock Tail assisted dynamic self righting.
\newblock In {\em Adaptive Mobile Robotics}. World Scientific, 2012.

\bibitem{kalakrishnan2009learning}
Mrinal Kalakrishnan, Jonas Buchli, Peter Pastor, and Stefan Schaal.
\newblock Learning locomotion over rough terrain using terrain templates.
\newblock In {\em IROS}, 2009.

\bibitem{keselman2017intel}
Leonid Keselman, John Iselin~Woodfill, Anders Grunnet-Jepsen, and Achintya
  Bhowmik.
\newblock Intel realsense stereoscopic depth cameras.
\newblock In {\em CVPR Workshops}, 2017.

\bibitem{khan2017memory}
Arbaaz Khan, Clark Zhang, Nikolay Atanasov, Konstantinos Karydis, Vijay Kumar,
  and Daniel~D. Lee.
\newblock Memory augmented control networks.
\newblock {\em arXiv:1709.05706}, 2017.

\bibitem{khatib1986potential}
Oussama Khatib.
\newblock The potential field approach and operational space formulation in
  robot control.
\newblock In {\em Adaptive and Learning Systems}. Springer, 1986.

\bibitem{khatib1986real}
Oussama Khatib.
\newblock Real-time obstacle avoidance for manipulators and mobile robots.
\newblock In {\em Autonomous robot vehicles}. Springer, 1986.

\bibitem{khoramshahi2013piecewise}
Mahdi Khoramshahi, Hamed~Jalaly Bidgoly, Soroosh Shafiee, Ali Asaei, Auke~Jan
  Ijspeert, and Majid~Nili Ahmadabadi.
\newblock Piecewise linear spine for speed--energy efficiency trade-off in
  quadruped robots.
\newblock {\em Robotics and Autonomous Systems}, 2013.

\bibitem{kim2020vision}
Donghyun Kim, D Carballo, Jared Di~Carlo, Benjamin Katz, Gerardo Bledt, Bryan
  Lim, and Sangbae Kim.
\newblock Vision aided dynamic exploration of unstructured terrain with a
  small-scale quadruped robot.
\newblock In {\em ICRA}, 2020.

\bibitem{kimmel1996fast}
R Kimmel and JA Sethian.
\newblock Fast marching methods for robotic navigation with constraints.
\newblock {\em Center for Pure and Applied Mathematics Report, University of
  California, Berkeley}, 1996.

\bibitem{kingma2014adam}
Diederik~P. Kingma and Jimmy Ba.
\newblock Adam: {A} method for stochastic optimization.
\newblock In {\em ICLR}, 2015.

\bibitem{koenig2005fast}
Sven Koenig and Maxim Likhachev.
\newblock Fast replanning for navigation in unknown terrain.
\newblock {\em IEEE Transactions on Robotics}, 2005.

\bibitem{kolter2008control}
J.~Zico Kolter, Mike~P Rodgers, and Andrew~Y. Ng.
\newblock A control architecture for quadruped locomotion over rough terrain.
\newblock In {\em ICRA}, 2008.

\bibitem{rma}
Ashish Kumar, Zipeng Fu, Deepak Pathak, and Jitendra Malik.
\newblock {RMA: Rapid Motor Adaptation for Legged Robots}.
\newblock In {\em RSS}, 2021.

\bibitem{kumar2018visual}
Ashish Kumar, Saurabh Gupta, David Fouhey, Sergey Levine, and Jitendra Malik.
\newblock Visual memory for robust path following.
\newblock In {\em NeurIPS}, 2018.

\bibitem{lavalle1998rapidly}
Steven~M LaValle.
\newblock Rapidly-exploring random trees: A new tool for path planning.
\newblock Technical report, Iowa State University, 1998.

\bibitem{lew2019contact}
Thomas Lew, Tomoki Emmei, David~D Fan, Tara Bartlett, Angel Santamaria-Navarro,
  Rohan Thakker, and Ali-akbar Agha-mohammadi.
\newblock Contact inertial odometry: collisions are your friends.
\newblock {\em arXiv preprint arXiv:1909.00079}, 2019.

\bibitem{li2021planning}
Tianyu Li, Roberto Calandra, Deepak Pathak, Yuandong Tian, Franziska Meier, and
  Akshara Rai.
\newblock Planning in learned latent action spaces for generalizable legged
  locomotion.
\newblock {\em RA-L}, 2021.

\bibitem{lillicrap2016continuous}
Timothy~P. Lillicrap, Jonathan~J. Hunt, Alexander Pritzel, Nicolas Heess, Tom
  Erez, Yuval Tassa, David Silver, and Daan Wierstra.
\newblock Continuous control with deep reinforcement learning.
\newblock In {\em ICLR}, 2016.

\bibitem{magana2019fast}
Octavio Antonio~Villarreal Magana, Victor Barasuol, Marco Camurri, Luca
  Franceschi, Michele Focchi, Massimiliano Pontil, Darwin~G Caldwell, and
  Claudio Semini.
\newblock Fast and continuous foothold adaptation for dynamic locomotion
  through cnns.
\newblock {\em RA-L}, 2019.

\bibitem{mastalli2017trajectory}
Carlos Mastalli, Michele Focchi, Ioannis Havoutis, Andreea Radulescu, Sylvain
  Calinon, Jonas Buchli, Darwin~G Caldwell, and Claudio Semini.
\newblock Trajectory and foothold optimization using low-dimensional models for
  rough terrain locomotion.
\newblock In {\em ICRA}, 2017.

\bibitem{mastalli2015line}
Carlos Mastalli, Ioannis Havoutis, Alexander~W Winkler, Darwin~G Caldwell, and
  Claudio Semini.
\newblock On-line and on-board planning and perception for quadrupedal
  locomotion.
\newblock In {\em 2015 IEEE International Conference on Technologies for
  Practical Robot Applications}, 2015.

\bibitem{mezghani2021memory}
Lina Mezghani, Sainbayar Sukhbaatar, Thibaut Lavril, Oleksandr Maksymets, Dhruv
  Batra, Piotr Bojanowski, and Karteek Alahari.
\newblock Memory-augmented reinforcement learning for image-goal navigation.
\newblock {\em arXiv:2101.05181}, 2021.

\bibitem{mirowski2018learning}
Piotr Mirowski, Matt Grimes, Mateusz Malinowski, Karl~Moritz Hermann, Keith
  Anderson, Denis Teplyashin, Karen Simonyan, Koray Kavukcuoglu, Andrew
  Zisserman, and Raia Hadsell.
\newblock Learning to navigate in cities without a map.
\newblock In {\em NeurIPS}, 2018.

\bibitem{miura1984dynamic}
Hirofumi Miura and Isao Shimoyama.
\newblock Dynamic walk of a biped.
\newblock {\em IJRR}, 1984.

\bibitem{mnih2016asynchronous}
Volodymyr Mnih, Adria~Puigdomenech Badia, Mehdi Mirza, Alex Graves, Timothy
  Lillicrap, Tim Harley, David Silver, and Koray Kavukcuoglu.
\newblock Asynchronous methods for deep reinforcement learning.
\newblock In {\em ICML}, 2016.

\bibitem{montemerlo2002fastslam}
Michael Montemerlo, Sebastian Thrun, Daphne Koller, and Ben Wegbreit.
\newblock Fastslam: A factored solution to the simultaneous localization and
  mapping problem.
\newblock 2002.

\bibitem{pyrobot2019}
Adithyavairavan Murali, Tao Chen, Kalyan~Vasudev Alwala, Dhiraj Gandhi, Lerrel
  Pinto, Saurabh Gupta, and Abhinav Gupta.
\newblock Pyrobot: An open-source robotics framework for research and
  benchmarking.
\newblock {\em arXiv:1906.08236}, 2019.

\bibitem{nachum2020multi}
Ofir Nachum, Michael Ahn, Hugo Ponte, Shixiang~Shane Gu, and Vikash Kumar.
\newblock Multi-agent manipulation via locomotion using hierarchical sim2real.
\newblock In {\em CoRL}, 2020.

\bibitem{parisotto2017neural}
Emilio Parisotto and Ruslan Salakhutdinov.
\newblock Neural map: Structured memory for deep reinforcement learning.
\newblock {\em arXiv:1702.08360}, 2017.

\bibitem{peng2018sim}
Xue~Bin Peng, Marcin Andrychowicz, Wojciech Zaremba, and Pieter Abbeel.
\newblock Sim-to-real transfer of robotic control with dynamics randomization.
\newblock In {\em ICRA}, 2018.

\bibitem{peng2020learning}
Xue~Bin Peng, Erwin Coumans, Tingnan Zhang, Tsang-Wei~Edward Lee, Jie Tan, and
  Sergey Levine.
\newblock Learning agile robotic locomotion skills by imitating animals.
\newblock In {\em RSS}, 2020.

\bibitem{raibert1984hopping}
Marc~H. Raibert.
\newblock Hopping in legged systems—modeling and simulation for the
  two-dimensional one-legged case.
\newblock {\em IEEE Transactions on Systems, Man, and Cybernetics}, 1984.

\bibitem{librealsense}
Intel RealSense.
\newblock {librealsense}.
\newblock \url{https://github.com/IntelRealSense/librealsense}.

\bibitem{savinov2018semi}
Nikolay Savinov, Alexey Dosovitskiy, and Vladlen Koltun.
\newblock Semi-parametric topological memory for navigation.
\newblock In {\em ICLR}, 2018.

\bibitem{habitat19iccv}
Manolis Savva, Abhishek Kadian, Oleksandr Maksymets, Yili Zhao, Erik Wijmans,
  Bhavana Jain, Julian Straub, Jia Liu, Vladlen Koltun, Jitendra Malik, Devi
  Parikh, and Dhruv Batra.
\newblock Habitat: {A} {P}latform for {E}mbodied {AI} {R}esearch.
\newblock In {\em ICCV}, 2019.

\bibitem{schulman2017proximal}
John Schulman, Filip Wolski, Prafulla Dhariwal, Alec Radford, and Oleg Klimov.
\newblock Proximal policy optimization algorithms.
\newblock {\em arXiv:1707.06347}, 2017.

\bibitem{sethian1999fast}
James~A Sethian.
\newblock Fast marching methods.
\newblock {\em SIAM review}, 1999.

\bibitem{smith2021legged}
Laura Smith, J~Chase Kew, Xue~Bin Peng, Sehoon Ha, Jie Tan, and Sergey Levine.
\newblock Legged robots that keep on learning: Fine-tuning locomotion policies
  in the real world.
\newblock In {\em ICRA}, 2022.

\bibitem{song2020rapidly}
Xingyou Song, Yuxiang Yang, Krzysztof Choromanski, Ken Caluwaerts, Wenbo Gao,
  Chelsea Finn, and Jie Tan.
\newblock Rapidly adaptable legged robots via evolutionary meta-learning.
\newblock In {\em IROS}, 2020.

\bibitem{sreenath2011compliant}
Koushil Sreenath, Hae-Won Park, Ioannis Poulakakis, and Jessy~W Grizzle.
\newblock A compliant hybrid zero dynamics controller for stable, efficient and
  fast bipedal walking on mabel.
\newblock {\em IJRR}, 2011.

\bibitem{stentz1997optimal}
Anthony Stentz.
\newblock Optimal and efficient path planning for partially known environments.
\newblock In {\em ICRA}, 1994.

\bibitem{szot2021habitat}
Andrew Szot, Alex Clegg, Eric Undersander, Erik Wijmans, Yili Zhao, John
  Turner, Noah Maestre, Mustafa Mukadam, Devendra Chaplot, Oleksandr Maksymets,
  Aaron Gokaslan, Vladimir Vondrus, Sameer Dharur, Franziska Meier, Wojciech
  Galuba, Angel Chang, Zsolt Kira, Vladlen Koltun, Jitendra Malik, Manolis
  Savva, and Dhruv Batra.
\newblock Habitat 2.0: Training home assistants to rearrange their habitat.
\newblock In {\em NeurIPS}, 2021.

\bibitem{tan2018sim}
Jie Tan, Tingnan Zhang, Erwin Coumans, Atil Iscen, Yunfei Bai, Danijar Hafner,
  Steven Bohez, and Vincent Vanhoucke.
\newblock Sim-to-real: Learning agile locomotion for quadruped robots.
\newblock In {\em RSS}, 2018.

\bibitem{thrun2002probabilistic}
Sebastian Thrun.
\newblock Probabilistic robotics.
\newblock {\em Communications of the ACM}, 2002.

\bibitem{tobin2017domain}
Josh Tobin, Rachel Fong, Alex Ray, Jonas Schneider, Wojciech Zaremba, and
  Pieter Abbeel.
\newblock Domain randomization for transferring deep neural networks from
  simulation to the real world.
\newblock In {\em IROS}, 2017.

\bibitem{truong2020learning}
Joanne Truong, Denis Yarats, Tianyu Li, Franziska Meier, Sonia Chernova, Dhruv
  Batra, and Akshara Rai.
\newblock Learning navigation skills for legged robots with learned robot
  embeddings.
\newblock {\em arXiv:2011.12255}, 2020.

\bibitem{wellhausen2021rough}
Lorenz Wellhausen and Marco Hutter.
\newblock Rough terrain navigation for legged robots using reachability
  planning and template learning.
\newblock In {\em IROS}, 2021.

\bibitem{wermelinger2016navigation}
Martin Wermelinger, P{\'e}ter Fankhauser, Remo Diethelm, Philipp Kr{\"u}si,
  Roland Siegwart, and Marco Hutter.
\newblock Navigation planning for legged robots in challenging terrain.
\newblock In {\em IROS}, 2016.

\bibitem{wijmans2019dd}
Erik Wijmans, Abhishek Kadian, Ari Morcos, Stefan Lee, Irfan Essa, Devi Parikh,
  Manolis Savva, and Dhruv Batra.
\newblock Dd-ppo: Learning near-perfect pointgoal navigators from 2.5 billion
  frames.
\newblock {\em arXiv:1911.00357}, 2019.

\bibitem{wooden2010autonomous}
David Wooden, Matthew Malchano, Kevin Blankespoor, Andrew Howardy, Alfred~A
  Rizzi, and Marc Raibert.
\newblock Autonomous navigation for bigdog.
\newblock In {\em ICRA}, 2010.

\bibitem{xia2018gibson}
Fei Xia, Amir~R Zamir, Zhiyang He, Alexander Sax, Jitendra Malik, and Silvio
  Savarese.
\newblock Gibson env: Real-world perception for embodied agents.
\newblock In {\em CVPR}, 2018.

\bibitem{xie2020dynamics}
Zhaoming Xie, Xingye Da, Michiel van~de Panne, Buck Babich, and Animesh Garg.
\newblock Dynamics randomization revisited: A case study for quadrupedal
  locomotion.
\newblock In {\em ICRA}, 2021.

\bibitem{yang2021real}
Bowen Yang, Lorenz Wellhausen, Takahiro Miki, Ming Liu, and Marco Hutter.
\newblock Real-time optimal navigation planning using learned motion costs.
\newblock In {\em ICRA}, 2021.

\bibitem{yang2022learning}
Ruihan Yang, Minghao Zhang, Nicklas Hansen, Huazhe Xu, and Xiaolong Wang.
\newblock Learning vision-guided quadrupedal locomotion end-to-end with
  cross-modal transformers.
\newblock In {\em ICLR}, 2022.

\bibitem{yang2021fast}
Yuxiang Yang, Tingnan Zhang, Erwin Coumans, Jie Tan, and Byron Boots.
\newblock Fast and efficient locomotion via learned gait transitions.
\newblock In {\em CoRL}, 2021.

\bibitem{yin2007simbicon}
KangKang Yin, Kevin Loken, and Michiel Van~de Panne.
\newblock Simbicon: Simple biped locomotion control.
\newblock {\em ACM Transactions on Graphics}, 2007.

\bibitem{yu2019biped}
Wenhao Yu, Visak C.~V. Kumar, Greg Turk, and C.~Karen Liu.
\newblock Sim-to-real transfer for biped locomotion.
\newblock In {\em IROS}, 2019.

\bibitem{yu2018policy}
Wenhao Yu, C.~Karen Liu, and Greg Turk.
\newblock Policy transfer with strategy optimization.
\newblock In {\em ICLR}, 2018.

\bibitem{yu2020learning}
Wenhao Yu, Jie Tan, Yunfei Bai, Erwin Coumans, and Sehoon Ha.
\newblock Learning fast adaptation with meta strategy optimization.
\newblock {\em RA-L}, 2020.

\bibitem{yu2017preparing}
Wenhao Yu, Jie Tan, C.~Karen Liu, and Greg Turk.
\newblock Preparing for the unknown: Learning a universal policy with online
  system identification.
\newblock In {\em RSS}, 2017.

\bibitem{zhang2017neural}
Jingwei Zhang, Lei Tai, Ming Liu, Joschka Boedecker, and Wolfram Burgard.
\newblock Neural slam: Learning to explore with external memory.
\newblock {\em arXiv:1706.09520}, 2017.

\bibitem{zhao2021surprising}
Xiaoming Zhao, Harsh Agrawal, Dhruv Batra, and Alexander~G Schwing.
\newblock The surprising effectiveness of visual odometry techniques for
  embodied pointgoal navigation.
\newblock In {\em ICCV}, 2021.

\bibitem{zhou2019environment}
Wenxuan Zhou, Lerrel Pinto, and Abhinav Gupta.
\newblock Environment probing interaction policies.
\newblock In {\em ICLR}, 2019.

\bibitem{zhu2017target}
Yuke Zhu, Roozbeh Mottaghi, Eric Kolve, Joseph~J Lim, Abhinav Gupta, Li
  Fei-Fei, and Ali Farhadi.
\newblock Target-driven visual navigation in indoor scenes using deep
  reinforcement learning.
\newblock In {\em ICRA}, 2017.

\end{thebibliography}
}

\newpage
\appendix
\section{Locomotion Policy Details}

\noindent\textbf{Base Policy \& Env-Factor Encoder Architecture:} We follow the implementation of~\cite{rma}. The base walking policy is a multi-layer perceptron (MLP) with 3 hidden layers. The input is the current state $x_t \in \mathbb{R}^{30}$, previous action $a_{t-1} \in \mathbb{R}^{12}$ and the extrinsics vector $z_t \in \mathbb{R}^{8}$ and the output is 12-dim target joint angles. The dimension of hidden layers is 128. The extrinsics vector $z_t$ is estimated by an environment factor encoder. The environment factor encoder is a 3-layer MLP (256, 128 hidden layer sizes) and encodes $e_t \in \mathbb{R}^{17}$ into $z_t \in \mathbb{R}^8$. 

\vspace{0.6em}\noindent\textbf{Adaptation Module Architecture:} The adaptation module first embeds states and actions into 32-dim vector using a 2-layer MLP. Then, a 3-layer $1$-D CNN convolves the representations across the time dimension to capture temporal correlations in the input. The input channel number, output channel number, kernel size, and stride of each layer are $[32, 32, 8, 4], [32, 32, 5, 1], [32, 32, 5, 1]$. The flattened CNN output is linearly projected to estimate $\hat{z_t}$. 

\vspace{0.6em}\noindent\textbf{Learning the Walking Policy:} We jointly train the base policy and the environment encoder network using PPO ~\cite{schulman2017proximal} for $15,000$ iterations ($1.2$B sample, ~24 hours) each of which uses batch size of $80,000$ split into $4$ mini-batches. We then train the adaptation module using supervised learning with on-policy data. We run the optimization process for $1000$ iterations ($80$M samples, ~3 hours) and use Adam optimizer ~\cite{kingma2014adam} to minimize MSE loss. The batch size is $80,000$ split up into $4$ mini-batches.

\vspace{0.6em}\noindent\textbf{Reward Function:} The reward at time $r_t$ is defined as the sum of the following quantities:
\begin{itemize}[nosep]
    \item Velocity Matching: $- | v_x  - v^{\mathrm{cmd}}_x | - |\omega_{\mathrm{yaw}} - \omega^{\mathrm{cmd}}_{\mathrm{yaw}}|$
    \item Energy Consumption: $- {\bm{\tau}^T \bm{\dot{q}}}$
    \item Lateral Movement: $- |v_y| ^ 2$
    \item Hip Joints: $- \|\bm{q}_{\mathrm{hip}}\| ^ 2$
\end{itemize}
The corresponding scalings are 20, 0.075, 1 and 0.2. The survival bonus is set by a simple rule as $10 + 20 (v^{\mathrm{cmd}}_x + \omega^{\mathrm{cmd}}_{\mathrm{yaw}})$. 

We list the ranges of command linear velocity and angular velocity in Supplementary Table~\ref{tab:vel-range}. We re-sample the command velocities within a single episode with probability $0.004$.   


\section{Safety Advisor Details} 

\noindent\textbf{Hyperparameters:} Velocity changes in Fall Predictor and the size of obstacles in Collision Detector are set by simple rules. For instance, (a) if a fall is predicted, the safety advisor module decreases the velocity limit by a large amount (we pick 0.2 m/s), so the robot can slow down quickly; (b) otherwise, it increases the velocity limit by a small amount (we pick 0.05 m/s) for conservative speed up; (c) the size of obstacles in Collision Detector (9cm x 3cm) is set to roughly be the size of the head of the robot. Additional real-world experiments [\href{https://youtu.be/OAEOZV76PcY}{link}] show that if the obstacle is set to be larger, the robot will take a more conservative path around the unexpected obstacle. If the obstacle is set to be smaller, the robot takes a shorter path but risks colliding legs with the unexpected obstacle.

\vspace{0.6em}\noindent\textbf{Network Structure:} Similar to the adaptation module, both the collision detector and fall predictor module share the same architecture and embed states and actions into a 32-dim vector using a linear layer. Then, we use 3 layers of 1D convolutions with input channels, output channels and strides $[32, 32, 8, 4], [32, 32, 5, 1], [32, 32, 5, 1]$. The output is a sigmoid scalar. 

\begin{table}[t]
\begin{center}
\resizebox{\linewidth}{!}{
\begin{tabular}{lcc}
\toprule
 \textbf{Task} & \textbf{\thead{Command Linear Velocity \\ Range (m / s)}} & \textbf{\thead{Command Angular Velocity \\ Range (rad / s)}}\\
 \midrule
 Curve Following & [0.15, 1.0] & [-0.4, 0.4]\\
 In-Place Turning & [0, 0.15] &  [-0.6, 0.6]\\
\bottomrule
\end{tabular}
}
\caption{Command velocity range for curve following and in-place turning.}
\label{tab:vel-range}
\vspace{-3em}
\end{center}
\end{table}

\vspace{0.6em}\noindent\textbf{Training Data and Environments:} The scalar sigmoid output predicts a probability value, indicating whether the robot collides with an obstacle in the Obstacle Detector, or if the robot falls at time $t + 100$ and $0$ otherwise (note that one simulation time-step is 0.01s) in the Fall Predictor. We train both modules in an self-supervised fashion by collecting data from robot walking / colliding with the obstacles / falling down. Data are collected given random command linear/angular velocity commands in environments with randomly sampled frictions, terrain roughness and payload values from the following list:
\begin{itemize}[nosep]
    \item Coefficient of Friction: $[0.1, 0.6, 1.1, 1.6, 2.1]$.
    \item Payload: $[1.2, 2.4, 3.6, 4.8, 6.0]$ (kg).
    \item Rough Terrain z-scale: $[0.01, 0.08, 0.14, 0.23]$ (m).
    \item Linear Velocity: $[0, 0.5, 1.0]$ (m/s).
    \item Angular Velocity: $[-0.4, 0.0, 0.4]$ (rad/s).
\end{itemize}
We train both Obstacle Detector and Fall Predictor for 145k iterations with a batch size of 1000. At simulation test time, we run both the collision detector and fall predictor at 5Hz whereas for deployment on robot we train a lightweight version using only the last 20 timesteps of observation history and run it at 10Hz.

\section{Visual Planner Details}
We command the angular velocity for our robot and the baseline LoCoBot using the following equation:
\begin{align}
\omega^{\mathrm{cmd}}_t = K_p \cdot (\theta^{\mathrm{target}}_t - \theta_t) + K_d \cdot (\omega^{\mathrm{target}}_t - \omega_t)
\end{align}
where $K_p = 1$, $K_d = 0.02$, $\omega^{\mathrm{target}}$ is set to 0. The command angular velocity is clipped to the range in Supplementary Table~\ref{tab:vel-range} before being sent to the locomotion policy in order to be consistent with the training setting. We also observe that when the linear speed is low (less than 0.1m/s), the locomotion policy is unable to make in-place turns with a small commanded angular velocity, due to the imperfection of our locomotion policy. Thus in this case we clip the absolute value of the commanded angular velocity to be at least 0.4 to compensate this imperfection. We empirically observe a higher performance even when the command is sub-optimal, mainly because our planning algorithm operates in a relatively high frequency and can soon correct the angular velocity command as soon as the linear velocity becomes large enough.

\section{LoCoBot Baseline \& \\Discrete Planner Details}

\begin{figure}[t]
\centering
\includegraphics[width=0.75\linewidth]{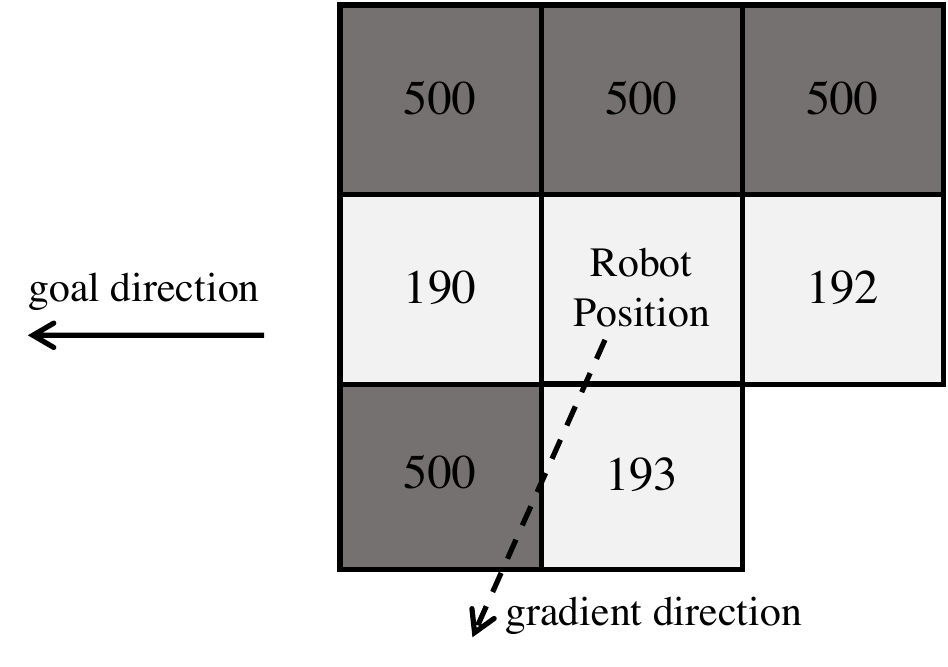}
\caption{An example of local minima produced by the map. The gray square represents the non-traversible areas. The white square represents the traversible regions. The number in each square represents the cost on that point. At the current position, the robot will orientes to the bottom left which the linear velocity commands will command 0 velocity, in which case the robot got stuck in the local minima.}
\vspace{-1em}
\label{fig:local_min}
\end{figure}

We import the PyRobot URDF model~\cite{pyrobot2019}. Both our method and the LoCoBot use a control frequency of 100Hz and a planning frequency of 10Hz. We follow~\cite{dudek2010computational} to convert commanded linear and angular velocity to the angular speed of the left and right wheel of the LoCoBot. We set the forward action of the discrete planner at 0.6 m/s after we measured the average speed of the continuous planner in the same evaluation environment is around 0.6 m/s. The low level controller is also a PD controller with $K_p = 10$, $K_d = 0.05$. The controller gain is adjusted so that no obvious motion jerk happens during movement. Since the control of wheeled robot is simpler and more accurate, we do observe the LoCoBot being more likely to stuck in local minima in the cost map (an illustration is shown in Supplementary Figure~\ref{fig:local_min}). For our robot, since the locomotion policy is not perfect and the legged robot is harder to control compared with LoCoBot, it sometimes can get out of the local minima due to the noisy movement, which is the reason why we perform better in the perfect flat ground (Table 4 (a) and (b) in the main text). However, we want to emphasize again that our point here is not to show our robot performs slightly better than baseline in the flat ground. Instead, what we show is the ability to traverse and navigate over difficult terrains where LoCoBot easily fail (Table 4 (c), (d), and (e) in the main text).

\end{document}